\documentclass[runningheads]{llncs}
\usepackage{graphicx}
\usepackage{acro}
\usepackage{amsmath}
\usepackage{amssymb}
\usepackage{graphicx}
\usepackage{times}
\usepackage{epsfig}
\usepackage{graphicx}
\usepackage{amsmath}
\usepackage{amssymb}
\usepackage{bm}
\usepackage{makecell}
\usepackage{mathtools}
\usepackage{pifont}
\usepackage[pagebackref=true,breaklinks=true,colorlinks,bookmarks=false]{hyperref}

\DeclareMathOperator*{\argmin}{arg\,min}
\DeclareMathOperator{\rep}{rep}
\DeclareMathOperator{\conv}{conv}
\DeclareAcronym{ALE}{
short=ADDLE,
long=auto-decoded deep latent embeddings
}

\DeclareAcronym{US}{
short=US,
long=ultrasound
}

\DeclareAcronym{JT}{
short=JT,
long=Jonckheere-Terpstra
}

\DeclareAcronym{BD}{
short=\textbf{BD},
long=big-data
}

\DeclareAcronym{CAD}{
short=CAD,
long=computer-aided diagnosis
}

\DeclareAcronym{AUC}{
short=AUC,
long=area under the curve,
long-plural-form=areas under the curve,
foreign-plural={}
}

\DeclareAcronym{ROC}{
short=ROC,
long=receiver operating characteristic
}

\DeclareAcronym{FPR}{
short=FPR,
long=false positive rate
}

\DeclareAcronym{EM}{
short=EM,
long=expectation-maximization
}

\DeclareAcronym{HP}{
short=\textbf{HP},
long=histopathology,
}

\DeclareAcronym{PCA}{
short=PCA,
long=principal components analysis,
}

\DeclareAcronym{PACS}{
short=PACS,
long=picture archiving and communication system
}

\newcommand{\etal}{\textit{et al.}}

\newcommand{\ie}{\textit{i.e.},}

\begin{document}

\title{Learning from Subjective Ratings Using Auto-Decoded Deep Latent Embeddings}
\titlerunning{Auto-Decoded Deep Latent Embeddings}
%

\author{Bowen Li\inst{1} \and Xinping Ren\inst{2} \and Ke Yan\inst{1} \and Le Lu\inst{1}\and Lingyun Huang\inst{4}\and Guotong Xie\inst{4} \and Jing Xiao\inst{4} \and Dar-In Tai\inst{3} \and Adam P. Harrison \inst{1}}


\authorrunning{B.Li et al.}

\institute{PAII Inc., Bethesda, MD 20817, USA \and
Ruijin Hospital, Shanghai, China
\and
Chang Gung Memorial Hospital, Linkou, Taiwan, ROC \and
PingAn Technology, Shenzhen, China}

 \authorrunning{B. Li et al.}
\acresetall
\maketitle              
\begin{abstract}
Depending on the application, radiological diagnoses can be associated with high inter- and intra-rater variabilities. Most \ac{CAD} solutions treat such data as incontrovertible, exposing learning algorithms to considerable and possibly contradictory label noise and biases. Thus, managing subjectivity in labels is a fundamental problem in medical imaging analysis. To address this challenge, we introduce  \ac{ALE}, which explicitly models the tendencies of each rater using an auto-decoder framework. After a simple linear transformation, the latent variables can be injected into any backbone at any and multiple points, allowing the model to account for rater-specific effects on the diagnosis. Importantly, \ac{ALE} does not expect multiple raters per image in training, meaning it can readily learn from data mined from hospital archives. Moreover, the complexity of training \ac{ALE} does not increase as more raters are added. During inference each rater can be simulated and a ``mean'' or ``greedy'' virtual rating can be produced. We test \ac{ALE} on the problem of liver steatosis diagnosis from 2D \ac{US} by collecting $36\,602$ studies along with clinical \ac{US} diagnoses originating from $65$ different raters. We evaluated diagnostic performance using a separate dataset with gold-standard \emph{biopsy} diagnoses. \ac{ALE} can improve the partial \acp{AUC} for diagnosing severe steatosis by $10.5\%$ over standard classifiers while outperforming other annotator-noise approaches, including those requiring $65$ times the parameters. 

\end{abstract}

\section{Introduction}

Deep learning has been widely applied to \acf{CAD} tasks~\cite{suzuki2017overview,cheng2021scalable}. In order to train effective deep neural networks, large-scale labeled datasets are needed~\cite{greenspan2016guest,Willemink_2020}. Yet, the labels within large-scale data, such as those found within hospital \acp{PACS}, are typically image-based diagnoses, which are not always considered ``ground-truth''~\cite{Willemink_2020}. Take for instance liver steatosis (fatty liver) diagnosis. Liver biopsy is considered the gold standard diagnosis for liver steatosis, but the risks and costs associated with such procedures hinder the collection of large-scale data. On the other hand, \acf{US} is the most common tool for assessing liver steatosis~\cite{Hernaez_2011} and image/US-diagnosis pairs can be readily mined from \acp{PACS}. Unfortunately, \ac{US} diagnoses are considered subjective and suffer from considerable inter- and intra-rater variability~\cite{Hernaez_2011}. Thus, \ac{CAD} solutions are faced with the quandary of needing to produce quantitative diagnostic scores when only subjectively labelled training data is available.

The problem of annotator noise has historically been addressed using \ac{EM} approaches~\cite{Dawid_1979,Welinder_2010,Khetan_2018}, which typically attempt to estimate true labels and a conditional model that explains rater labels. However, apart from Khetan \etal{}~\cite{Khetan_2018}, these require multiple ratings per sample, which is not a realistic requirement for large-scale data, especially for clinical \ac{PACS} data. Another approach for annotator noise is multi-rater consensus modeling (MRCM)~\cite{yu2020difficulty}, which relies on a  consensus loss that is only meaningful in a multiple rating per sample scenario.  STAPLE is another popular \ac{EM} approach that is used to impute ground truth from noisy labels~\cite{Warfield_2004}, but it has the same multiple-rater requirements. In our problem setting and most clinical scenarios, there is only one rater annotating one image. Our solution is able to work with multi-rater data, but multi-rater algorithms cannot work with our data. In terms of methods that do not assume multi-ratings per training sample, Tanno \etal{} proposed an approach to estimate annotator confusion matrices that avoids the complexity of \ac{EM}~\cite{Tanno_2019}, but they  assume that any annotator noise is dependent only on the true label and independent of the image itself, which will likely be violated. Another common  limitation of the above approaches is that they are only applicable for categorical classification.  More akin to our approach, some recent investigations have explicitly modelled each individual rater. For instance Chou \etal{} train separate models for each individual or group of raters~\cite{Chou_2019}, but this is very computationally expensive. Guan \etal{} train a separate classification head for each rater~\cite{Guan_2018}, but this is limited in modelling capability and their final solution requires a weighted average that again relies on multiple raters per sample, which are needed to calculate weights. 

To fill these gaps, we introduce \acf{ALE}. In contrast to these above approaches, \ac{ALE} models individual raters using a rich and expressive latent embedding that is probabilistically motivated. After a linear transformation to match dimensions, the encodings in rater-specific latent vectors can be added to convolutional or global features. The latent vector values, along with shared backbone weights, that best predict each rater's labels are then learned. \ac{ALE} avoids the complexity of \ac{EM} approaches and can be readily trained using typical gradient descent procedures. Additionally, although it admits multiple raters per sample, \ac{ALE} naturally works with single ratings per sample, addressing the most critical use-case within \ac{CAD}. Moreover, \ac{ALE} correctly models subjective ratings as conditional on both the image itself and the true label. Finally, \ac{ALE} imposes no restrictions on what loss can be used and can be readily applied to any medical imaging task, including classification, detection, or segmentation.

We validate \ac{ALE} on the task of liver steatosis diagnosis using conventional liver \ac{US} images. Liver steatosis affects $20-30\%$ of the global population and is associated with serious risk factors, such as liver fibrosis and hepatocellular carcinoma~\cite{Younossi_2018}. We trained \ac{ALE} on a large-scale multi-scanner and multi-etiology \ac{PACS}-mined dataset of image/\ac{US} diagnosis pairs ($3\,790$ patients, $312\,848$ images, and $36\,602$ studies) that were labelled by $65$ different clinicians during clinical care. Apart from our methodological contributions, this alone represents a significant contribution in its own right. Deep learning has previously been applied to liver \ac{US}, such as for diagnosing steatosis~\cite{byra2018transfer,biswas2018symtosis,gummadi2020automated,reddy2018novel}, diagnosing fibrosis~\cite{li2020reliable}, and detecting lesions~\cite{wu2014deep}. For liver steatosis, datasets are small, with less than $200$ patients and less than $1000$ images~\cite{byra2018transfer,biswas2018symtosis,gummadi2020automated,reddy2018novel} with only single scanners and etiologies, so method generalizability is under-tested. 

In terms of methodological contributions, we show that \ac{ALE} can learn from and model subjective \ac{US} diagnoses. We evaluated \ac{ALE} on a separate dataset of $148$ patients with biopsy-proven diagnoses and show that the ``mean'' and ``greedy'' \ac{ALE} virtual raters can outperform both standard classification baselines and also leading annotator noise approaches~\cite{Guan_2018,Chou_2019} in discriminating between healthy, mild, moderate, and severe steatosis. These results demonstrate that \ac{ALE} provides a flexible, straightforward, and effective approach to manage subjectivity in medical imaging labels.

\section{Methods} \label{sec:method}
\begin{figure*}[t]
\includegraphics[width=\textwidth]{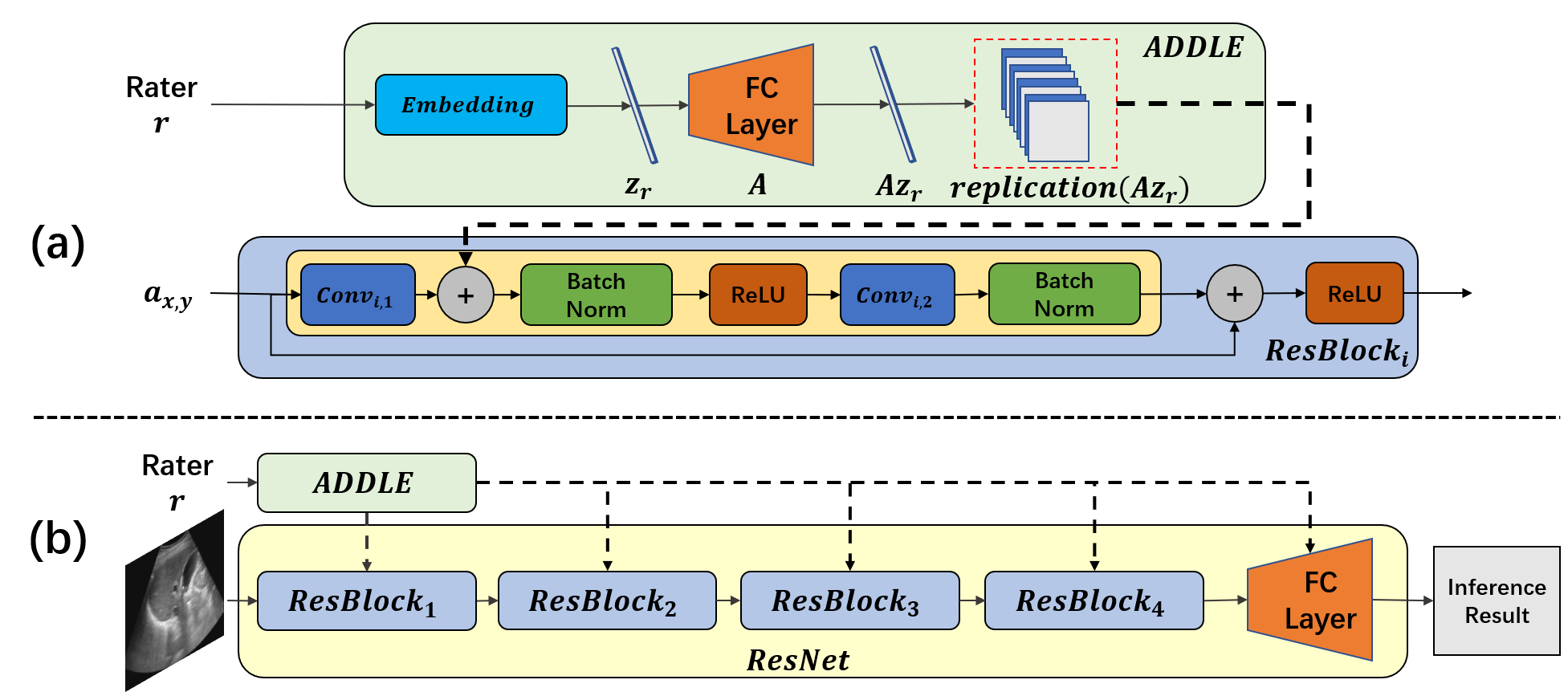}
\caption{Algorithmic workflow of \acs{ALE}. (a): \acs{ALE} and its integration into a ResBlock. (b): Potential injection points of \acs{ALE} into a ResNet. \acs{ALE} can also be used with any other classifcation, detection, or segmentation backbone.}
\label{fig-alg} 
\end{figure*}

Fig.~\ref{fig-alg} depicts the workings of \ac{ALE}. In short, we assume we are given a dataset of images, noisy/subjective labels, and rater indices  $\mathcal{X}=\{\mathbf{x}_{i},y_{i},\,r\}_{i=1}^{N}$ where $r=1,\ldots R$ indexes which rater generated the label. Here for simplicity we assume a single rater per image and image-level labels, but \ac{ALE} can easily admit multiple raters and other label types. We describe the training and inference process below. 

\subsubsection{Training}

Park \etal{}~\cite{park2019deepsdf} introduced an auto-decoder-based method, which learns the latent space of shapes for shape representation. \ac{ALE} uses a similar approach to create rater-specific latent embeddings. We associate a set of latent codes, $\mathcal{Z}=\{\mathbf{z}_{r}\}_{r=1}^{R}$, to the raters. In practical terms, each $\mathbf{z}_{r}$ is a simple vector. We model the mapping from images and latent embeddings to a subjective rating using a deep network, $f_{\theta}(\mathbf{x}_{i}, \mathbf{z}_{r})$, that accepts both the image and the corresponding latent code as input. The deep network can be any popular or well-known network that would normally accept the $\mathbf{x}_{i}$ input. We can inject the latent codes to any convolutational or global feature map along the network's forward propagation path via simple addition. In this way, the shared neural net weights encode common features that can be influenced by the latent embeddings to model any rater-specific tendencies. To retain any pretrained weights, we only inject the latent codes as part of an existing convolutional or fully-connected layer. For instance, if the latent code dimensionality is $M$ and we wish to inject the latent codes in conjunction with a convolution with $C$ outputs operating on a feature map, $\mathbf{a}_{x,y}$, we simply apply a linear transformation and replication operation:
\begin{align}
    \mathbf{\tilde{a}}_{x,y} = \conv(\mathbf{a}_{x,y}) + \rep(\mathbf{A}\mathbf{z}_{r}) \textrm{,} \label{eqn:injection}
\end{align}
where $\mathbf{\tilde{a}}_{x,y}$ is the output feature map, $\conv(.)$ is the convolutional operator, $\mathbf{A}$ is an $C\times M$ matrix, and $\rep(.)$ replicates a vector across the convolutional spatial dimensions. Due to the linearity of convolution, \eqref{eqn:injection} operates as if a large convolution was applied to a concatenated feature representation of $\mathbf{a}_{x,y}$ and $\mathbf{z}_{r}$. Importantly, any original pretrained weights in $\conv(.)$ can be kept. If injected to a global feature, the $\rep(.)$ operator can be forgone. Finally, if beneficial \ac{ALE} allows latent codes to be injected at multiple points. 

To learn the latent codes, we formulate a posterior composed of a product between a prior distribution and the likelihood of the observed labels:
\begin{align}
    p_{\theta}(\mathcal{Z}|\mathcal{X})=\prod_{r}p(\mathbf{z}_{r})\prod_{\{\mathbf{x}_{i},y_{i}\}\in \mathcal{X}_{r}}p_{\theta}(y_{i}|\mathbf{x}_{i}, \mathbf{z}_{r}) \textrm{,} \label{eqn:full_prob}
\end{align}
where $\mathcal{X}_{r}$ selects all samples labelled by the $r$th rater and $\theta$ parameterizes the likelihood. We assume that the prior, $p(\mathbf{z}_{r})$, follows an isotropic zero-mean Gaussian distribution with a covariance of $\sigma^{2}\mathbf{I}$ (which is also how we initialize the latent embeddings). Given any loss, $\mathcal{L}(.,.)$, the likelihood can be expressed using the deep neural network described above: 
\begin{align}
   p_{\theta}(y_{i}|\mathbf{x}_{i}, \mathbf{z}_{r}) = \exp(-\mathcal{L}(f_{\theta}(\mathbf{x}_{i}, \mathbf{z}_{r}), y_{i})) \textrm{.}
\end{align}
Loss functions are not constrained to be classification-based---indeed our own experiments employ an ordinal regression formulation. During training, \eqref{eqn:full_prob} is maximized by minimizing the following sum with respect to the rater codes, $\mathcal{Z}$, and the shared network parameters, $\theta$:
\begin{align}
    \argmin_{\theta,\,\mathcal{Z}}\sum_{\{\mathbf{x}_{i},y_{i},\,r\}\in\mathcal{X}}\left(\mathcal{L}(f_{\theta}(\mathbf{x}_{i}, \mathbf{z}_{r}), y_{i})\right) + \sum_{r=1}^{R}\left(\dfrac{1}{\sigma^2}||\mathbf{z}_{r}||_{2}^{2}\right) \textrm{,} \label{eqn:loss}
\end{align}
which can be trained using gradient descent. The auto-decoding nature of \ac{ALE} comes from the fact there is no encoding function for the latent embeddings. Instead the latent embedding values are learned solely based on the loss and prior formulation in \eqref{eqn:loss}. One downside of \eqref{eqn:loss} is that if there are unbalanced numbers of samples labeled by the raters, then poorly represented raters may not be well optimized. To deal with this, after a solution to \eqref{eqn:loss} is found the shared weights can be frozen and each rater's latent embedding can be individually fine-tuned:
\begin{align}
    \argmin_{\mathbf{z}_{r}}\sum_{\{\mathbf{x}_{i},y_{i}\}\in\mathcal{X}_{r}}\left(\mathcal{L}(f_{\theta}(\mathbf{x}_{i}, \mathbf{z}_{r}), y_{i})\right) + \dfrac{1}{\sigma^2}||\mathbf{z}_{r}||_{2}^{2} \textrm{,} \label{eqn:finetune}
\end{align}
where only the rater's samples, $\mathcal{X}_{r}$, are selected. Optimizing \eqref{eqn:finetune} is quick, since only each rater's latent vector needs to be fine-tuned, which are small in dimension.

\subsubsection{Inference}

After training, \ac{ALE} should model how each rater would label a new image. But, emulating rater subjectivity alone is not necessarily useful in inference. Because \ac{ALE} can provide simulated predictions for each rater, if there is \emph{a priori} information on which raters are more experienced or trustworthy, \ac{ALE} could simply simulate those ones. More generally, this is not available. One option is to greedily choose the ``best'' raters to average using a validation set with gold-standard diagnoses:
\begin{align}
    \bar{y} = \dfrac{1}{R} \sum_{r \in \mathcal{R}_{\mathrm{greedy}}}f_{\theta}(\mathbf{x}_{i}, \mathbf{z}_{r}) \textrm{,} \label{eqn:greedy}
\end{align}
where $\mathcal{R}_{\mathrm{greedy}}$ is a  set of raters that are greedily chosen until validation performance tops out. When such validation sets are not available \ac{ALE} can output a mean or majority rating: 
\begin{align}
    \bar{y} = \dfrac{1}{R} \sum_{r=1}^{R}f_{\theta}(\mathbf{x}_{i}, \mathbf{z}_{r}) \textrm{.} \label{eqn:mean_rater}
\end{align}

\section{Experiments}

\subsubsection{Datasets}

We test \ac{ALE} on the problem of liver steatosis diagnosis from \ac{US}.  Because \ac{US} is the most common modality for clinically assessing liver steatosis, it is possible to collect large-scale datasets for algorithmic training.  To this end, we collected a \ac{BD} cohort consisting of $3\,790$ patients, $312\,848$ \ac{US} images, and $36\,602$ \ac{US} studies from the \ac{PACS} of \emph{Anonymized}.  Through the course of clinical care, each study was given a four-class ordinal assessment of either healthy, mild, moderate, or severe steatosis from one of $65$ clinicians. We used $3\,405$ patients for training and left the rest as a stopping criteria validation set. We evaluated whether the \ac{ALE} greedy and mean raters of \eqref{eqn:greedy} and \eqref{eqn:mean_rater}, respectively, can provide a better quantitative score to differentiate more objective histopathology-derived steatosis severities. To do this we collected a separate \ac{HP} dataset of $218$ patients and \ac{US} studies, all with accompanying biopsy-proven diagnoses within $3$ months of the scan date. Histopathological diagnoses follow the same ordinal configuration from healthy to severe steatosis as the \ac{US} ones. We used $70$ of the patients for validation (\ac{HP}-V), but only for model selection and greedy rater selection, and \emph{not} as a validation set for stopping criteria. This left $148$ patients as a test set (\ac{HP}-T), with $25$, $35$, $36$, and $52$ diagnosed with healthy, mild, moderate, and severe steatosis, respectively . Patients with hepatitis B, hepatitis C, and non-alcoholic fatty liver disease are all represented in \ac{BD} and \ac{HP} studies originated from three different scanners with images corresponding to the six different viewpoints described by Li \etal{}~\cite{li2020reliable}.

\subsubsection{Setup}

 The well-known binary decomposition loss of Frank and Hall for ordinal regression~\cite{frank_simple_2001,furnkranz_binary_2009} was used for training on the ordered \ac{US} severity labels. We used a ResNet-18~\cite{he2016deep} backbone, which proved the most optimal, choosing a latent embedding dimensionality of $10$, injecting the latent codes at the beginning of the second residual block, and using a $\sigma^2$ value of $1.0$. An analysis of hyper-parameter sensitivity, found in the supplementary, indicates that performance was not sensitive to the choice of $\sigma^2$ and was stable across most injection points and latent code sizes. Once trained, a simple summation of the Frank and Hall outputs can produce a single score~\cite{furnkranz_binary_2009}. In training we treat each \ac{US} image in a study as an independent sample, but in inference we take the mean score across all images in a study to produce a study-wise score. Analysis on \ac{HP}-V indicated that the top-two raters should be used for the greedy \ac{ALE} variant of \eqref{eqn:greedy}.

As baselines, we compare \ac{ALE} to both a standard ResNet-18 network trained on \ac{BD}'s \ac{US} labels (denoted ResNet-18-\ac{BD}) and a ResNet-18 trained only on the small-scale \ac{HP}-T dataset using five-fold cross validation (denoted ResNet-18-\ac{HP}). Comparisons with these baselines respectively reveal the impact of modelling rater tendencies and the importance of using large-scale data for training, even should it be subjectively labelled. We also compared against annotator noise methods that permit ordinal regression. The first comparison uses Chou \etal{}'s JLSL approach of training a separate model for each rater~\cite{Chou_2019}. Note, Chou \etal{} also used additional components, but these rely on multiple raters per sample or the availability of gold-standard labels during training (which we do not assume), so we only evaluate the idea of a single model per rater. We also evaluate Guan \etal{}'s multi-head approach of using a separate classification layer for each rater~\cite{Guan_2018}, but forego their weighting process that also assumes multiple raters per sample.
In short, JLSL~\cite{Chou_2019} trains a separate model for each rater and multi-head~\cite{Guan_2018} trains a separate classification head. Because we test on the same backbone, their implementation details are identical to ADDLE’s apart from the above critical differences. For fairness, we finetune multi-head using an analogous version of \eqref{eqn:finetune}. Like \ac{ALE}, for each we evaluate their ``mean'' and ``greedy'' rater performance.

\subsubsection{Evaluation Protocols}

Because the problem is to discriminate between four ordered histopathological grades, standard binary \ac{ROC} analysis cannot be performed. Following standard practice, we use \ac{ROC} analysis on three different and ordered binary \emph{cutoffs}, specifically $>=$ mild, $>=$ moderate, or $=$ severe levels of steatosis. Associated \ac{AUC} summary statistics for these cutoffs are denoted $AUC_0$, $AUC_1$, and $AUC_2$, respectively. Like was done by Li \etal{} for liver fibrosis~\cite{li2020reliable}, we primarily focus on partial \ac{ROC} curves and \acp{AUC} only within the range of false positive rates that are $<=30\%$. The reasoning being that lower specificities are not clinically useful operating points to investigate. Partial \acp{AUC} are normalized to be within a range of $0$ to $1$. Additionally, we also report \ac{JT} index values~\cite{jonckheere1954distribution}, which is a multi-partite generalization of the \ac{AUC}~\cite{furnkranz_binary_2009} that also ranges from $0$ to $1$, with $1$ corresponding to perfect discrimination between the four histopathological grades.

\subsubsection{Results}

\begin{table*}[h]
\caption{Diagnostic performance on the biopsy-proven test set. Note, ResNet18-\acs{HP} is cross-validated on the \ac{HP}-T test set itself, whereas all others are trained on the large-scale \ac{BD} dataset with subjective \ac{US} labels. }
\begin{center}
\begin{tabular}{c|c|cc|cc|cc|c}
\Xhline{2\arrayrulewidth}

 &  &\multicolumn{2}{c|}{$>=$ Mild}&\multicolumn{2}{c|}{$>=$ Moderate}&\multicolumn{2}{c|}{$=$ Severe} & Params.\\

\textbf{Model} &  $JT$ &$AUC_0$& $AUC_{0P}$ &  $AUC_1$ & $AUC_{1P}$ &$AUC_2$&$AUC_{2P}$ & (millions)\\

\Xhline{2\arrayrulewidth}

ResNet-18-\acs{BD}  & $0.882$&$0.952$&$0.876$&$0.923$&$0.770$&$0.869$&$0.615$ & $11.00$ \\

ResNet-18-\acs{HP} & $0.816$&$0.799$&\textbf{}$0.509$&$0.872$&$0.642$&$0.892$&$0.676$ & $11.00$\\
\Xhline{\arrayrulewidth}
JLSL~\cite{Chou_2019} (mean) & $0.892$&$0.956$&$0.869$&$0.916$&$0.736$&$0.893$&$0.697$& $714.95$\\

JLSL~\cite{Chou_2019} (greedy) &
$0.878$&$0.944$&$0.860$&$0.897$&$0.722$&$0.883$&$0.694$ & $714.95$\\
\Xhline{\arrayrulewidth}
Multi-head~\cite{Guan_2018} (mean) &$0.888$&$0.956$&$\mathbf{0.875}$&$0.918$&$0.746$&$0.879$&$0.658$ & $11.03$\\

Multi-head~\cite{Guan_2018} (greedy) & $0.889$&$0.955$&$0.869$&$0.924$&$0.764$&$0.873$&$0.650$ & $11.03$\\
\Xhline{\arrayrulewidth}

\acs{ALE} (mean) &$0.890$&$\mathbf{0.957}$&$0.873$&$0.922$&$0.762$&$0.886$&$0.672$ & $11.00$\\
\acs{ALE} (greedy) &$\mathbf{0.898}$&$0.955$&$0.864$&$\mathbf{0.931}$&$\mathbf{0.786}$&$\mathbf{0.899}$&$\mathbf{0.720}$ & $11.00$\\

\Xhline{2\arrayrulewidth}
\end{tabular}
\end{center}
\label{table-performance-test}
\end{table*}

\begin{figure}[t]
\includegraphics[width=\textwidth]{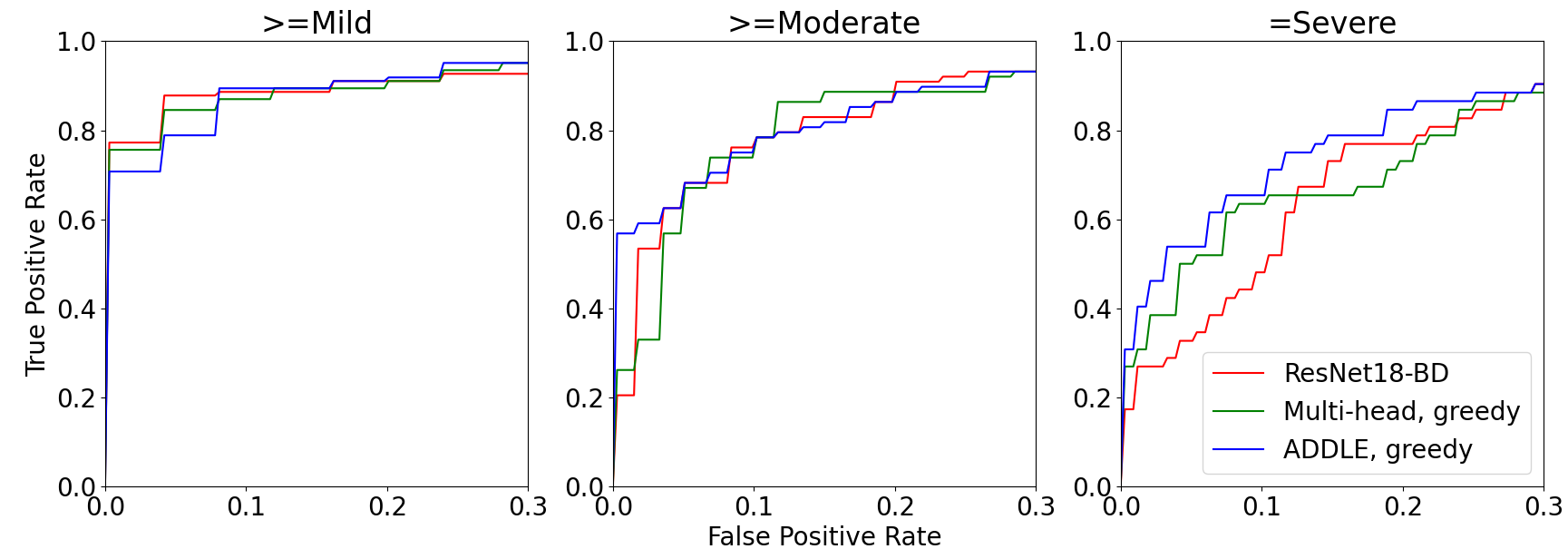}\textbf{}
\caption{Partial ROC curves (\acsp{FPR} $\leq0.30$) on the biopsy-proven test set, excluding JLSL which requires $65$ models.}
\label{fig:roc} 
\end{figure}

Table~\ref{table-performance-test} outlines the performance of all tested models on \ac{HP}-T, and Fig.~\ref{fig:roc} depicts selected partial \ac{ROC} curves of methods that only require one model. 
As can be seen, the standard ResNet-18 trained on \ac{US} labels can post a good JT score and good \acp{AUC}. However, it struggles to identify severe steatosis, posting quite poor partial \acp{AUC}. As Fig.~\ref{fig:roc}(c) demonstrates, this corresponds to only a sensitivity of $40\%$ at a \ac{FPR} of $10\%$. Overall, ResNet-18-\ac{HP} performs much poorer, even though it was cross-validated on \ac{HP}-T itself and trained with actual histopatholgoical labels. This highlights the importance of training on large-scale data even if rater noise is present. Moving on to the competitor models, they are mostly able to boost the \ac{JT} scores. For the most part, any gains are seen in the $AUC_2$ scores, with JLSL outperforming multi-head. However, for JLSL these gains require training $65$ separate models. Moreover, only greedy multi-head is able to match the baseline's $AUC_1$ scores. Finally, JLSL's greedy rater performance is considerably worse than its mean rater variant, suggesting an overfitting problem that prevents rater performances generalizing from \ac{HP}-V to \ac{HP}-T. 

When \ac{ALE}'s result are examined, it can be seen that it can boost the $AUC_2$ by even greater margins, with the greedy rater performing best (partial $AUC_{2}$ increases from $0.615$ to $0.720$). Compared to ResNet-18-\ac{BD}, this is a boost of sensitivity from $40\%$ to $65\%$ at a $10\%$ \ac{FPR}. Greedy \ac{ALE} can also boost the $AUC_1$ results (partial \acp{AUC} increase from $0.770$ to $0.786$).  Importantly, despite only requiring one model and incurring no practical computational  cost over the baseline model, greedy \ac{ALE} can still outperform the much costlier greedy JLSL. When greedy rater selection is not available, \ac{ALE} and JLSL are more comparable, but mean \ac{ALE} achieves this performance at $1/65$ of the training cost. 

\begin{figure}[t]
\includegraphics[width=\textwidth]{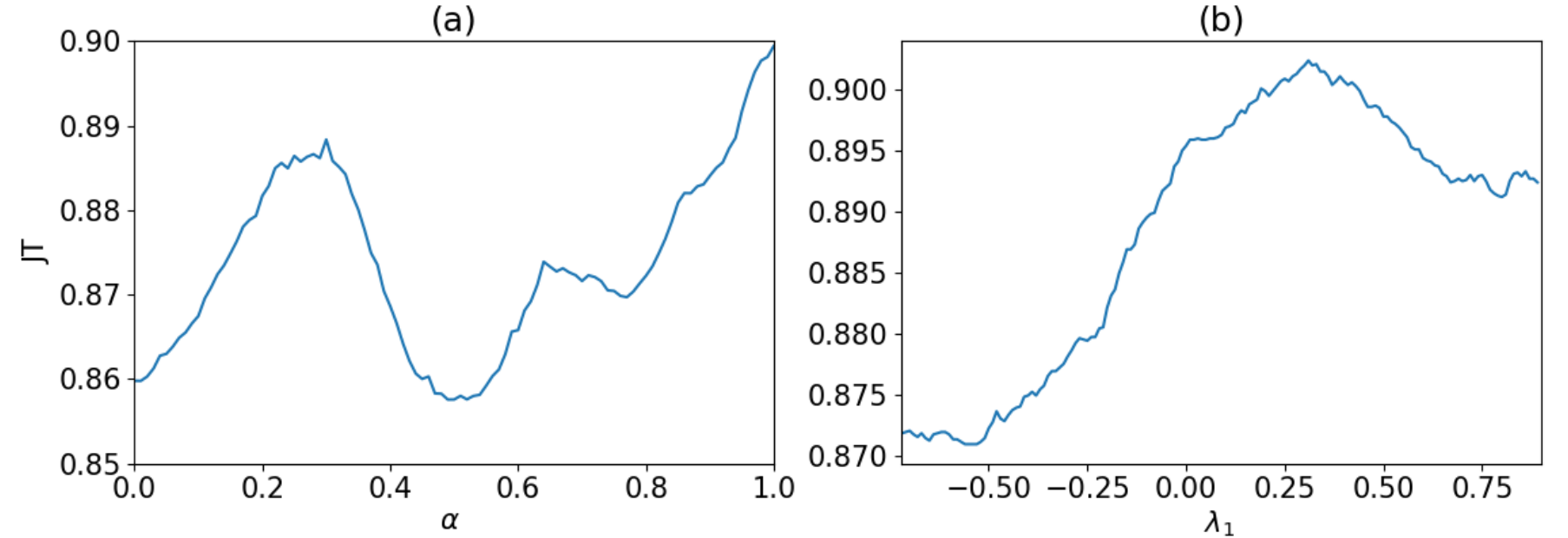}

\caption{The latent space and the effects on performance. (a): the test-set \ac{JT} scores are measured as the latent vector is interploated between the virtual raters with the worst ($\mathbf{z}_0$) and best ($\mathbf{z}_1$) performance, \ie{} $\mathbf{z}=\mathbf{z}_0+\alpha\cdot(\mathbf{z}_1-\mathbf{z}_0)$. (b): the performance when using the first principle component of the \acs{PCA} basis, scaled using  $\lambda_{1}=-0.72,\ldots 0.90$ (range calculated by projecting virtual raters to the first principle component). Graphs of other principle components can be found in the supplementary.}
\label{fig:latent-space} 
\end{figure}

The \ac{ALE} latent space can itself be analyzed, which is an avenue of analysis not available elsewhere. For instance, as Fig.~\ref{fig:latent-space}(a) demonstrates, there is considerable difference between the worst and best performing virtual-rater performance. Interpolating between the corresponding latent vectors produces a smooth performance curve. Additionally, as Fig.~\ref{fig:latent-space}(b) illustrates, when conducting \ac{PCA}, varying the first principle component will produce variations in performance, suggesting that the latent embedding is indeed capturing differences in rater abilities. The  first 6 principle components explain $75.2\%$ variance, and the last principle component still explains $4.5\%$ variance. These evidences support our choosing of latent embedding dimensionality $10$, which is not likely to be further compressed based on the explained variance ratio. The norms of all latent vectors, $z$, range from $0.40$ to $1.86$, which is a reasonable range for our choice of $\sigma^2=1.0$. However, one interesting fact we observed is that 2-D \ac{PCA} and 2-D t-SNE plots show no clustering of latent vectors with high performance, which could be a future research direction, and further investigation may provide additional insights into the latent space properties.

\section{Conclusion}

We introduced \ac{ALE} as an effective approach to deal with subjective ratings using auto-decoded latent variables. \ac{ALE} is highly \emph{expressive}, \ie{} modelling rater labels as conditional on the image, \emph{flexible}, \ie{} admitting any loss and number of ratings per sample, and \emph{efficient}, \ie{} incurring no additional computational cost in training over standard models. We train \ac{ALE} on  $36\,602$ subjectively labelled \ac{US} studies for liver steatosis, by far the largest such dataset used to date. When evaluated on a separate biopsy-proven dataset, \ac{ALE} outperforms standard classifiers as well as leading annotator noise competitors~\cite{Guan_2018,Chou_2019}. These results indicate that \ac{ALE} can better learn from and exploit subjective labels to produce quantitative steatosis assessments. Given the prevalence of subjective labels in \ac{CAD} training data, future work should validate \ac{ALE} in other end applications, including for detection and segmentation tasks.

\bibliographystyle{splncs04}
\bibliography{refs}

\end{document}